%% file: samplepaper.tex
\documentclass[runningheads]{llncs}
\usepackage[T1]{fontenc}
\usepackage{multirow}
\usepackage{etoolbox}
\usepackage{amsmath}
\usepackage{graphicx}
\usepackage{subcaption}
\usepackage{caption}
\usepackage{booktabs} 

\usepackage{algorithm}
\usepackage{algorithmic}
\usepackage{amsmath}
\usepackage{amssymb}
\usepackage{float}
\usepackage{graphicx}
\usepackage{subcaption}

\usepackage{academicons}
\usepackage{xurl}
\usepackage{marvosym}
\usepackage{color}
\begin{document}
\title{ComFairGNN: Community Fair Graph Neural Network}
%
\author{Yonas Sium\textsuperscript{\Letter} \and Qi Li\orcidID{0000-0002-3136-2157}}

\authorrunning{F. Author et al.}
%
\institute{Iowa State University\\
\email{\{yas, qli\}@iastate.edu}}

\maketitle              
\begin{abstract}
Graph Neural Networks (GNNs) have become the leading approach for addressing graph analytical problems in various real-world scenarios. However, GNNs may produce biased predictions against certain demographic subgroups due to node attributes and neighbors surrounding a node. Understanding the potential evaluation paradoxes due to the complicated nature of the graph structure is crucial for developing effective GNN debiasing mechanisms. In this paper, we examine the effectiveness of current GNN debiasing methods in terms of unfairness evaluation. Specifically, we introduce a community-level strategy to measure bias in GNNs and evaluate debiasing methods at this level. Further, We introduce ComFairGNN, a novel framework designed to mitigate community-level bias in GNNs. Our approach employs a learnable coreset-based debiasing function that addresses bias arising from diverse local neighborhood distributions during GNNs neighborhood aggregation. Comprehensive evaluations on three benchmark datasets demonstrate our model's effectiveness in both accuracy and fairness metrics.

\keywords{fairness  \and graph neural network \and bias \and evaluation metrics}
\end{abstract}

\input{body/introduction}

\input{body/preliminaries}

\input{body/proposed_framework}
\input{body/experiments}

\input{body/related_work}

\input{body/conclusion}
\bibliographystyle{splncs04}
\bibliography{sample-base}
\end{document}

%% file: body/introduction.tex
\section{Introduction}

In today's interconnected world, graph learning supports various real-world applications, such as social networks, recommender systems , and knowledge graphs \cite{bourigault2014learning,nickel2015review}. Graph Neural Networks (GNNs) are powerful for graph representation learning and are used in tasks like node classification and link prediction by aggregating neighboring node information. However, GNNs often overlook fairness, leading to biased decisions due to structural bias and attribute bias like gender, race, and political ideology. These bias can cause ethical dilemmas in critical contexts, such as job candidate evaluations, where a candidate might be favored due to shared ethnic background or mutual acquaintances.
\begin{figure}
\begin{center}
\vspace{0.2in}
  \includegraphics[width=\columnwidth]{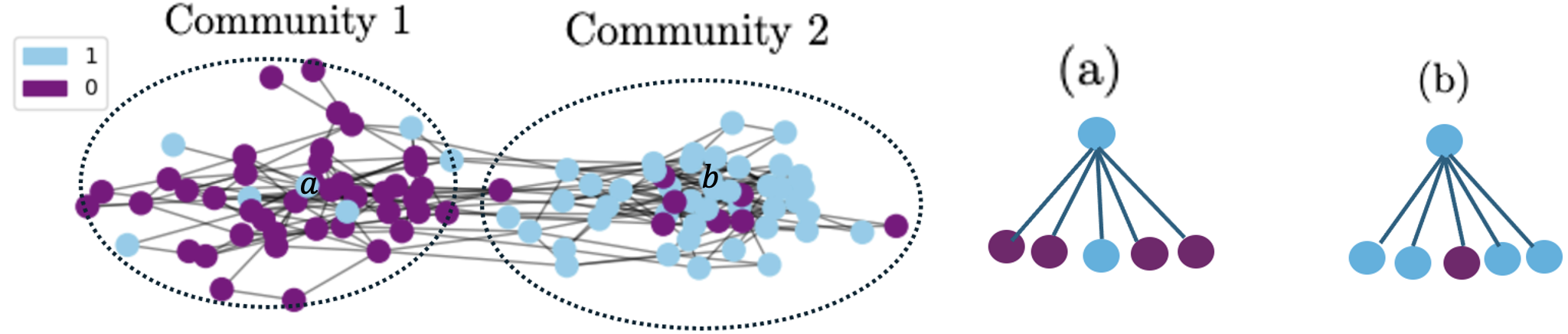}  
\end{center}
\vspace{-0.2in}
\caption{Illustration of graph polarization: Nodes $a$ and $b$ from different communities share identical labels (same colors), but GNNs learn different embeddings due to varying neighborhood label distributions.}
\label{fig:polarization}
\end{figure}
To tackle the outlined issue, multiple methods were suggested to evaluate and mitigate the fairness of node representation learning on graphs. Most of these methods aim to learn node representations that can yield statistically fair predictions across subgroups defined based on sensitive attributes. Choosing the right metric to evaluate bias in graphs inherently depends on the specific task at hand. The existing methods employed graph-level metrics originally designed for other purposes to address fairness concerns. The primary difficulty in addressing fairness within graphs arises from the frequent correlation between the graph's topology and the sensitive attribute we aim to disregard. However, due to the complicated nature of graph structures, conducting graph-level fairness evaluations and comparisons of different methods is not as straightforward as commonly reported in the existing methods.

Real-world graphs often polarize into communities based on sensitive attributes (e.g., age, gender, race, ideology), as shown in Figure \ref{fig:polarization}, where communities form through dense internal connections and sparse external links.  For example, U.S. Twitter networks show distinct political segregation \cite{guerra2013measure}, where Democratic supporters dominate Northeast and West Coast interactions while Republican supporters influence Texas, Florida, and Iowa networks. Similarly, sports preferences on social media can reflect regional demographic differences.

In this paper, we investigate the critical issue of community fairness in graph neural networks (GNNs) within the context of node classification. Simply using node attributes and structural information as a debiasing mechanism can lead to misleading fairness evaluations and loss of important information \cite{wu2019net}. Communities, which reflect local neighborhood structures, often stem from sensitive node attributes \cite{rahman2019fairwalk}. Specifically, GNNs generate node representations by aggregating information from neighboring nodes, which can result in nodes with identical labels but differing neighborhood label distributions, as illustrated in Figure~\ref{fig:polarization}. This leads to structural biases in the learning process. Moreover, such biases may be further amplified through multi-layered recursive neighborhood aggregation in GNNs. While current debiasing methods primarily focus on attribute-based fairness, they fail to address the underlying bias introduced by neighborhood aggregation. As a result, these methods are ineffective at mitigating community-level biases that arise from disparate neighborhood structures.

To address community fairness, we introduce the concept of local structural fairness at the community level for all nodes in the graph. This approach recognizes that the broader structural bias affecting nodes within a community emerges from the GNN's aggregation process, which is influenced by the diverse distribution of local neighborhoods. We present ComFairGNN, a novel community-fair Graph Neural Network compatible with any neighborhood aggregation based GNN architecture. Specifically, the coreset-based debiasing function targets the neighborhood aggregation of GNN operation, aiming to bring identical label nodes in different communities closer in the embedding space. This approach mitigates structural disparities across communities, leading to fairer representations for nodes with identical labels.

In summary, our paper makes three key contributions: (1) We measure the problem of fairness and potential evaluation paradoxes at the community level using different fairness evaluation metrics in GNNs. (2) We propose a novel community fairness for GNNs named ComFairGNN that can balance the structural bias for identical label nodes in different communities. ComFairGNN works with any neighborhood aggregation-based GNNs. (3) Comprehensive empirical evaluations validate our method's effectiveness in enhancing both fairness and accuracy.


%% file: body/preliminaries.tex
\section{Preliminaries}

We first introduce the notations. A graph $G = (V, E)$ consists of $n$ nodes and edges $E \subseteq V \times V$. The node-wise feature matrix $X \in \mathbb{R}^{n \times k}$ has $k$ dimensions for raw node features, with each row $\mathbf{x}_i$ representing the feature vector for the $i$-th node. The binary adjacency matrix is $A \in \{0, 1\}^{n \times n}$, and the learned node representations are captured in $H \in \mathbb{R}^{n \times d}$, where $d$ is the latent dimension size and $\mathbf{h}_i$ is the representation for the $i$-th node. The sensitive attribute $s \in \{0, 1\}^n$ classifies nodes into demographic groups, with edges $e_{uv}$ being intragroup if $s_u = s_v$ and intergroup otherwise. The function $N(u)$ returns the set of neighbors for a node $u$, providing a structural node embedding. Clusters based on this embedding are denoted as $C$, with $C_i \subseteq V$ for nodes in the $i$-th cluster and $c$ as the cluster count. Each $C_i$ represents a community and $f$ measures the statistical notion of fairness across different communities in the graph.

\subsection{Graph Neural Network}
\label{sec:gnn}
Graph Neural Networks (GNNs) utilize neighborhood aggregation across multiple layers. At layer $l$, the representation of node $v$, denoted as $\mathbf{h}_v^l \in \mathbb{R}^{d_l}$, is:
\begin{equation}
\mathbf{h}_v^l = \sigma \left(\text{AGGR} \left(\{\mathbf{h}_u^{l-1} \mid u \in \mathcal{N}_v\}; \omega^l\right)\right)
\label{eq:GNN}
\end{equation}
where $\text{AGGR}(\cdot)$ is an aggregation function like mean-pooling \cite{kipf2016semi} or self-attention \cite{velivckovic2017graph}, $\sigma$ is an activation function, $\mathcal{N}_v$ is $v$'s neighborhood set, and $\omega^l$ contains learnable parameters. The initial representation is $\mathbf{h}_v^0 = \mathbf{x}_v$, where $\mathbf{x}_v$ is the input feature vector.




\subsection{Node Clustering on Structural Embedding}
We partition the graph into communities using node2vec \cite{grover2016node2vec} embeddings combined with $k$-means clustering \cite{macqueen1967some}. Node2vec maps each node $u \in U$ to a $d$-dimensional vector space $f: U \rightarrow \mathbb{R}^d$. For each node, multiple random walks are performed to capture the network neighborhood $N_s(u)$, maximizing the objective:
\begin{equation}
\arg\max_{f} \prod_{u \in U} \prod_{u' \in N_s(u)} P(u' \mid f(u))
\end{equation}
where $P(u' \mid f(u))$ is computed using softmax. These structural embeddings are then clustered using $k$-means, chosen for its computational efficiency. The $k$-means algorithm minimizes the within-cluster sum of squares \cite{shiao2023carl}:
\begin{equation}
\arg\min \sum_{i=1}^{C}{\sum_{x \in C_{i}} \text{DIST}(x, \mu_i)}
\end{equation}
where $C_i$ represents the $i$-th cluster and $\mu_i$ its centroid.

%% file: body/proposed_framework.tex
\section{Proposed Method}

\begin{figure*}[t]
    \centering
    \includegraphics[width=\textwidth]{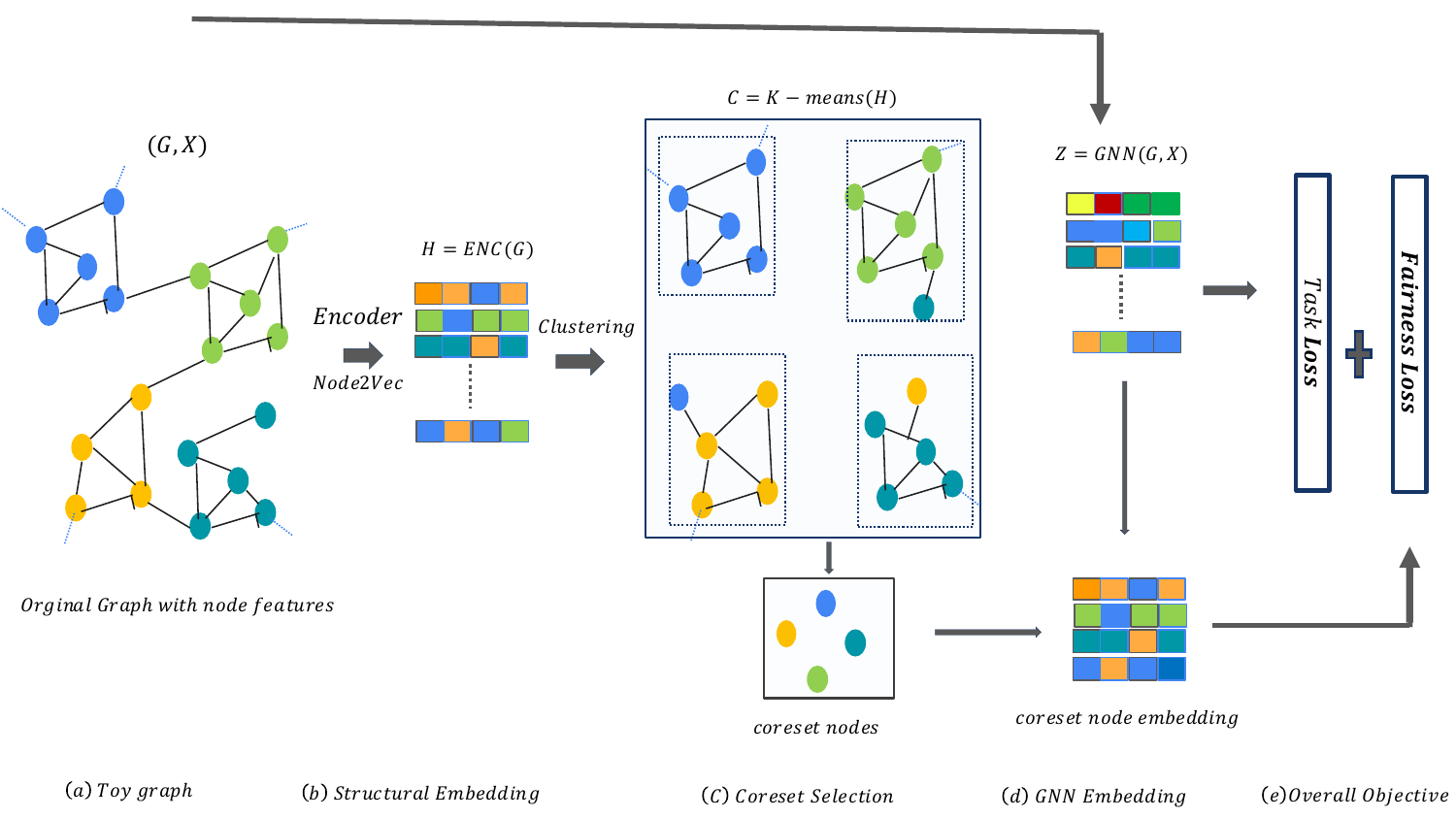}
    \caption{Overall framework of ComFairGNN}
    \label{fig:model}
\end{figure*}

Figure~\ref{fig:model} illustrates the ComFairGNN framework. We first identify communities using structural-based clustering (Fig.~\ref{fig:model}(c)). For debiasing, we divide each community's nodes into two subgroups based on sensitive features and sample coreset nodes according to neighborhood homophily ratios. These coreset nodes contrast against each other to mitigate structural bias (Fig.~\ref{fig:model}(d)). The GNN optimizes both task performance and coreset fairness (Fig.~\ref{fig:model}(e)), debiasing neighborhood aggregation by maximizing similarity between same-labeled nodes.

\subsection{Community Level Structural Contrast}


We propose a debiasing function using coreset nodes $c$ selected from different communities. Following \cite{toneva2018empirical}, coresets identify crucial training examples using heuristic criteria. We select nodes equally from subgroups $\mathcal{S}_0$ and $\mathcal{S}_1$ across communities. Since GNNs in Eq.~\ref{eq:GNN} only access one-hop contexts and communities tend to be homogeneous in sensitive attributes, selecting coreset nodes from both groups per community helps balance structural disparities.

\begin{equation}
C = \bigcup_{k = 1}^K C_{k_{S_0}} \cup \bigcup_{k = 1}^K C_{k_{S_1}} \subseteq V
\label{eq:cluster_group union}
\end{equation}

Here $C$ indicates the coreset sample nodes for all $k \in \{1, \ldots, K\}$. $K$ represents the total number of communities, and $C_{k_{S_0}}$ and $C_{k_{S_1}}$ represent selected sample nodes from subgroup $S_0$ and $S_1$ in community $k$ respectively.

To minimize structural bias between groups across communities, we maximize the similarity of similarly-labeled nodes in the coreset, allowing GNNs to implicitly reduce community-based structural differences.

\subsection{Coreset Nodes Selection for GNNs Debiasing}
\label{sec:homophily_ratio}
To fundamentally eliminate the structural bias of GNNs neighborhood aggregation, which is the key operation in GNNs, we propose to use the neighborhood homophily distribution ratio to select the sample coreset nodes from different communities. 

\noindent\textbf{Node neighborhood homophily distribution ratio:}
\label{subsec:neighbor_distribution}
The homophily ratio in graphs is typically defined based on the similarity between connected node pairs. In this context, two nodes are considered similar if they share the same node label. The formal definition of the homophily ratio is derived from this intuition, as follows \cite{ma2021homophily}. 

\begin{definition}[Node Homophily Ratio]
For a graph  $G = (\mathcal{V}, \mathcal{E})$ with node label vector $y$, we define the edge homophily ratio as the proportion of edges linking nodes sharing identical labels. The formal definition is as follows:
\begin{equation}
h(G, \{y_i; i \in \mathcal{V}\}) = \frac{1}{|n|} \sum_{(i,k)\in \mathcal{E}} f(y_i = y_k),
\end{equation}
where $|n|$ is the number of neighboring edges of node $i$ in the graph and $f(\cdot)$ is the indicator function.
\end{definition}
A node in a graph is generally classified as highly homophilous when its edge homophily ratio $h(\cdot)$ is high, typically falling within the range $0.5 \leq h(\cdot) \leq 1$, assuming an appropriate label context. Conversely, a node is considered heterophilous when its edge homophily ratio is low. Here, $h(\cdot)$ denotes the edge homophily ratio as previously defined.

\begin{figure}
    \centering
    \includegraphics [height=4cm, width=\columnwidth]{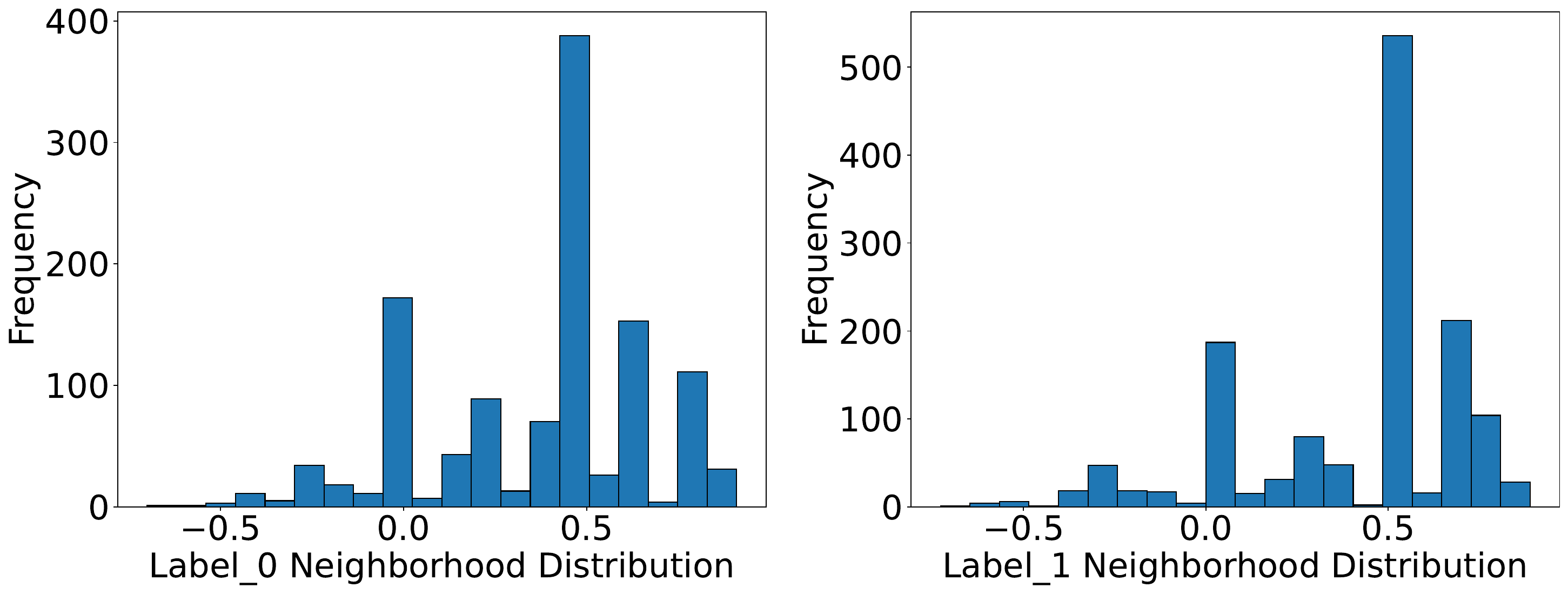}
    \caption{Edge distribution bias in Pokec-z: GNN embeddings vary for identical nodes based on neighborhood labels}
    \label{fig:edge_distr}
\end{figure}

Figure~\ref{fig:edge_distr} shows significant variation in neighborhood distribution ratios among same-labeled nodes in pokec-z. Since nodes with identical labels typically share similar features, GNN's 1-hop aggregation produces different embeddings for these nodes. This embedding disparity across communities may bias the GNN. Therefore, we sample debiasing coreset nodes from all communities based on their neighborhood distributions.

\textbf{Fairness-aware Graph Coreset:} We aim to identify a representative subset of training nodes $V_w \subset V_{\text{train}} = [n_t]$, where $|V_w| \leq c \ll n_t$, along with an associated set of sample weights. In nearest neighbor message passing models (such as GCN, GAT, and GIN), node $i$ only interacts with nodes $j$ that are within a shortest-path distance of $D$, denoted as $\{j \in [n] \mid d(i,j) < D\}$. Here, we assume node $i$ neighborhood distribution ratio section~\ref{subsec:neighbor_distribution}, to select the coreset nodes. This selection is designed to approximate the training loss $\mathcal{L}(\theta)$ across all $\theta \in \Theta$, while maintaining fairness considerations. The graph coreset selection problem is formulated as:

\begin{equation}
\min_{w \in \mathcal{W}} \max_{\theta \in \Theta} \left| \sum_{i \in [n_t]} w_i \cdot \ell([f_\theta(A,X)]_i, y_i) - \mathcal{L}(\theta) \right|
\end{equation}


The coreset minimizes the maximum error across all $\theta$, requiring both $A$ and $X$ matrices to compute $f_\theta(A, X)$. As illustrated in algorithm \ref{alg:fair_coreset}, to debias GNN neighborhood aggregation at the community level, we sample from both groups $S_0$ and $S_1$ across all communities.

\begin{algorithm}[H]
\caption{Fair Graph Coreset Selection}
\label{alg:fair_coreset}
\begin{algorithmic}[1]
\REQUIRE $G=(V,E)$, features $X$, groups $S_0$, $S_1$, size $K$ 
\ENSURE Coreset $C \subseteq V$ \COMMENT{Selected nodes}
\STATE Initialize $C = \emptyset$, $H = \text{ENC}(G)$ \COMMENT{Initial embedding}
\FOR{each community $k$ in $G$}
   \STATE $n_k = \lfloor K \cdot |V_k|/|V| \rfloor$ \COMMENT{Community budget}
   \STATE $C_{0k} = \text{Sample}(V_k \cap S_0, \lfloor n_k/2 \rfloor)$ 
   \STATE $C_{1k} = \text{Sample}(V_k \cap S_1, \lfloor n_k/2 \rfloor)$ 
   \STATE $C = C \cup C_{0k} \cup C_{1k}$ 
\ENDFOR
\STATE $Z = \text{PRD}(H)$ 
\STATE Optimize $C$ to minimize Task\_Loss + Fairness\_Loss 
\RETURN $C$
\end{algorithmic}
\end{algorithm}



\subsection{Training  Objective and Constraints:}
We focus on the task of node classification. We incorporate both the classification loss and the fairness loss associated with the classification to enhance the training process as illustrated in figure \ref{fig:model}(e).

\noindent\textbf{Classification loss:} In node classification tasks, the Graph Neural Network (GNNs) final layer typically adapts its dimensionality to align with the number of classes. It utilizes a softmax activation function, wherein the probability of class $i$ is represented by the $i^{th}$ output dimension. To optimize the model's parameters, we minimize the cross-entropy loss, defined as:

\begin{equation}
   \mathcal{L}_{task} = -\frac{1}{|Y|} \sum_{i} \sum_{j} Y_{i,j} \log(\hat{Y}_{i,j}) 
\end{equation}

Here, $|Y|$ denotes the number of labeled examples (rows in $Y$), while $\hat{Y}_{i,j}$ represents the predicted probability of the $i^{th}$ example belonging to the $j^{th}$ class.

\noindent\textbf{Fairness loss:}
 As the proposed coreset selection aims to maintain fair representations across different groups, we introduce a fairness loss on the selected coreset samples $c$. We group the samples in $c$ based on their class label. Let $c_{y=0}$ and $c_{y=1}$ denote the groups of samples in the coreset with labels $0$ and $1$, respectively for binary classification. We use a similarity-based loss, trying to achieve parity in the average pairwise similarities for the two label groups in $c$, as follows:

\begin{equation}
\mathcal{L}_{fair} = \left| \frac{1}{|c_{y=0}|} \sum\limits_{i,j \in c_{y=0}} \text{sim}(x_i, x_j) - \frac{1}{|c_{y=1}|} \sum\limits_{i,j \in c_{y=1}} \text{sim}(x_i, x_j) \right|
\end{equation}

where $\text{sim}(\cdot,\cdot)$ is a cosine similarity function between two samples of same label. This fairness loss drives the coreset selection toward balancing the average intra-label similarities between the two label groups in $c$. It aims to constrain the similarity distribution of labels to be similar across the two groups, $S_0$ and $S_1$.

\noindent\textbf{Overall loss:}
By combining all the above loss terms, we formulate the overall loss as

\begin{equation}
\mathcal{L} = \mathcal{L}_{\text{task}} + \mathcal{L}_{\text{fair}}
\end{equation}






%% file: body/experiments.tex
\section{Experiments}

We aim to address the following research questions in our experiments. \textbf{RQ1:} How reliable is the reported fairness score of graph debiasing methods free from paradox at a community level? \textbf{RQ2:} How efficient are the existing graph fairness evaluation methods in estimating the bias of GNNs at a community level? \textbf{RQ3:} How efficient is our proposed ComFairGNN method in terms of both accuracy and fairness at the community level?

\begin{table}[htbp]
\centering
\resizebox{1.0\columnwidth}{!}{
\begin{tabular}{|c|c|c|c|c|c|c|c|}
\hline\hline
Method & Dataset & ACC$\uparrow$ & AUC$\uparrow$ & $\Delta$EO(w)$\downarrow$ & $\Delta$SP(w)$\downarrow$ & $\Delta$EO (w/o) & $\Delta$SP (w/o) \\
\hline\hline
\multirow{3}{*}{Vanilla GCN} & Pokec-z & 69.29 $\pm$ 1.72 & 69.51 $\pm$ 2.7 & 14.43 $\pm$ 0.7 & 12.88 $\pm$ 0.94 & -14.43 $\pm$ 0.7 & -12.88 $\pm$ 0.94\\
& Pokec-n & \textbf{70.77 $\pm$ 1.32} & 70.45 $\pm$ 1.7 & 18.03 $\pm$ 0.57 & 14.14 $\pm$ 2.01 & 18.03 $\pm$ 0.57 & 14.14 $\pm$ 2.01 \\
& Facebook & 81.53 $\pm$ 1.52 & 61.62 $\pm$ 2.1 & 12.73 $\pm$ 0.7 & 12.30 $\pm$ 1.0 & -12.73 $\pm$ 0.7 & -12.30 $\pm$ 1.0 \\
\hline\hline
\multirow{3}{*}{FairGNN} & Pokec-z & 68.95 $\pm$ 0.1 & 72.24 $\pm$ 0.3 & 5.17 $\pm$ 1.3 & 7.79 $\pm$ 0.7 & -5.17 $\pm$ 1.3 & -6.79 $\pm$ 0.7 \\
& Pokec-n & 69.14 $\pm$ 0.1 & 70.18 $\pm$ 0.3 & 4.26 $\pm$ 1.3 & 3.87 $\pm$ 0.7 & 4.26 $\pm$ 1.3 & 3.87 $\pm$ 0.7 \\
& Facebook & 85.35 $\pm$ 0.77 & 64.62 $\pm$ 1.6 & \textbf{2.22 $\pm$ 0.6} & \textbf{6.36 $\pm$ 1.0} & 2.22 $\pm$ 0.6 & 6.36 $\pm$ 1.0 \\
\hline\hline
\multirow{3}{*}{NIFTY} & Pokec-z & 67.34 $\pm$ 1.51 & 67.49 $\pm$ 1.20 & 5.89 $\pm$ 1.32 & 5.85 $\pm$ 0.9 & -5.89 $\pm$ 1.32 & -5.85 $\pm$ 0.9 \\
& Pokec-n & 69.41 $\pm$ 1.51 & 69.08 $\pm$ 1.20 & 8.63 $\pm$ 1.32 & 10.42 $\pm$ 0.9 & -8.63 $\pm$ 1.32 & -10.42 $\pm$ 0.9 \\
& Facebook & 77.07 $\pm$ 1.28 & 73.05 $\pm$ 1.87 & 7.88 $\pm$ 1.0 & 8.95 $\pm$ 1.3 & 7.88 $\pm$ 1.0 & 8.95 $\pm$ 1.3 \\
\hline\hline
\multirow{3}{*}{UGE} & Pokec-z & 63.64 $\pm$ 2.57 & 64.61 $\pm$ 2.1 & 14.34 $\pm$ 2.3 & 11.74 $\pm$ 1.9 & -14.34 $\pm$ 2.3 & -11.74 $\pm$ 1.9 \\
& Pokec-n & 67.48 $\pm$ 2.72 & 66.67 $\pm$ 2.71 & 10.34 $\pm$ 2.3 & 12.74 $\pm$ 1.9 & -10.34 $\pm$ 2.3 & -12.74 $\pm$ 1.9 \\
& Facebook & 64.33 $\pm$ 0.94 & 63.46 $\pm$ 1.8 & 5.69 $\pm$ 1.2 & 13.17 $\pm$ 1.5 & 5.69 $\pm$ 1.2 & 13.17 $\pm$ 1.5 \\
\hline\hline
\multirow{3}{*}{ComFairGNN} & Pokec-z & \textbf{70.25 $\pm$ 0.1} & 71.11 $\pm$ 0.3 & \textbf{0.18 $\pm$ 1.3} & \textbf{1.00 $\pm$ 0.7} & 0.18 $\pm$ 1.3 & 1.00 $\pm$ 0.7 \\
& Pokec-n & 69.41 $\pm$ 0.7 & 66.43 $\pm$ 0.97 & \textbf{0.19 $\pm$ 0.45} & \textbf{0.90 $\pm$ 0.67} & -0.19 $\pm$ 0.45 & -0.90 $\pm$ 0.67 \\
& Facebook & \textbf{85.35 $\pm$ 0.77} & 74.62 $\pm$ 2.6 & 3.96 $\pm$ 0.6 & 7.63 $\pm$ 1.0 & 3.96 $\pm$ 0.6 & 7.63 $\pm$ 1.0 \\
\hline\hline
\end{tabular}
}
\caption{\small Comparison of GNNs on utility and bias at the graph level with different debiasing methods.}
\label{tab:results}
\end{table}

\begin{table}[h]
\centering
\resizebox{1.0\columnwidth}{!}{
\begin{tabular}{|l|r|r|r|r|c|c|c|c|c|}
\hline
Dataset & Nodes & Edges & Features & Classes & Comm\_1 & Comm\_2 & Comm\_3 & Comm\_4 & Comm\_5 \\
\hline
Pokec-z & 67,796 & 882,765 & 276 & 2 & 309 & 282 & 259 & 1412 & 302 \\
Pokec-n & 66,569 & 198,353 & 265 & 2 & 428 & 360 & 361 & 389 & - \\
Facebook & 1,034 & 26,749 & 573 & 2 & 55 & 37 & 65 & - & - \\
\hline
\end{tabular}
}
\caption{\small Dataset statistics and community distribution}
\label{tab:dataset}
\end{table}

\subsection{Experimental Setup and Datasets}
\noindent\textbf{Datasets.} We evaluate on three real-world datasets for node classification: \textit{Pokec-z}, \textit{Pokec-n}, and \textit{Facebook}. The Pokec datasets are subsets from Slovakia's Pokec social network \cite{takac2012data}, where nodes represent users, edges represent friendships, and location is the sensitive attribute for classifying users' work fields. The Facebook dataset \cite{he2016ups} similarly uses friendship connections with gender as the sensitive attribute. Dataset statistics are summarized in Table~\ref{tab:dataset}.

\noindent\textbf{Evaluation Metrics}
We use ACC and AUC scores to evaluate the utility performance. To evaluate fairness, we use two commonly used fairness metrics, i.e., $\Delta_{DP} = |P(\hat{y} = 1|s = 0) - P(\hat{y} = 1|s = 1)|$ \cite{dwork2012fairness} and $\Delta_{EO} = |P(\hat{y} = 1|y = 1, s = 0) - P(\hat{y} = 1|y = 1, s = 1)|$ \cite{hardt2016equality}. $\hat{y}$ and $y$ denote the node label prediction and ground truth, respectively. For $\Delta_{DP}$ and $\Delta_{EO}$, a smaller value indicates a fairer model prediction.


\noindent\textbf{Baseline Methods:} We investigate the effectiveness of four state-of-the-art GNN debiasing methods at a graph level and community level, namely FairGNN \cite{dai2021say}, NIFTY \cite{agarwal2021towards}, and UGE \cite{wang2022unbiased}. FairGNN utilizes adversarial training to eliminate sensitive attribute information from node embeddings. NIFTY improves fairness by aligning predictions based on both perturbed and unperturbed sensitive attributes. This approach generates graph counterfactuals by inverting the sensitive feature values for all nodes while maintaining all other attributes. UGE derives node embeddings from a neutral graph that is free of biases and unaffected by sensitive node attributes, addressing the issue of unbiased graph embedding. We use GCN \cite{kipf2016semi} as the backbone GNN model. 


\noindent\textbf{Implementation Setup:} 
We use a two-layer GCN \cite{kipf2016semi} encoder and MLP predictor for all experiments, consistent with baseline methods implemented from PyGDebias \cite{dong2023fairness}. Each method runs for 400 epochs with learning rates in \{0.1,0.01,0.001\}, using hyperparameters from their original papers and repeating three times. Experiments were conducted on 16GB NVIDIA V100 GPUs.


\captionsetup{font=small}

\begin{figure*}[t]
    \centering
    \begin{minipage}[t]{0.48\textwidth}
        \centering
        \includegraphics[width=\textwidth]{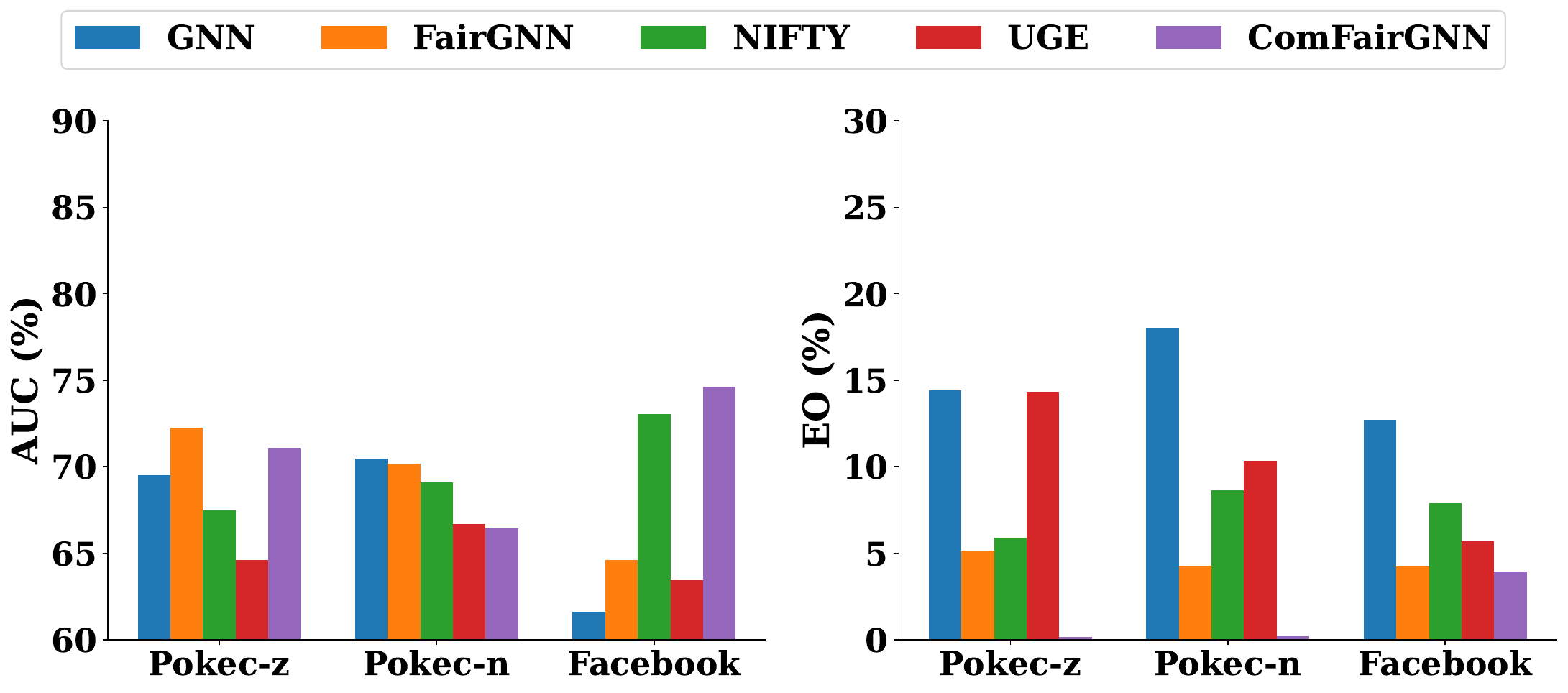}
        \caption*{(a) Performance at a graph level}
        \label{fig:dataset1}
    \end{minipage}
    \hfill
    \begin{minipage}[t]{0.48\textwidth}
        \centering
        \includegraphics[width=\textwidth]{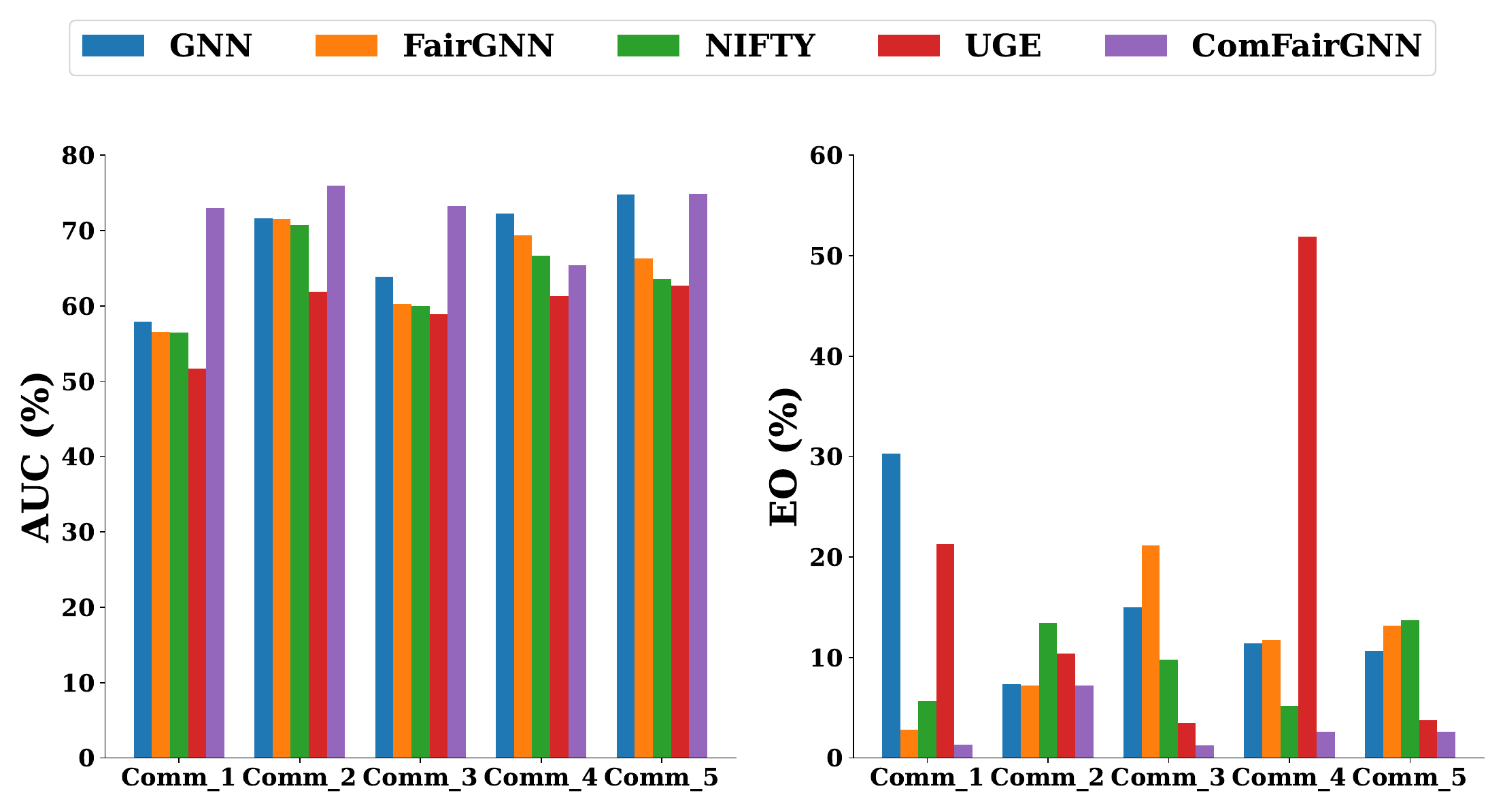}
        \caption*{ (b) Performance on Pokec-z communities}
        \label{fig:dataset2}
    \end{minipage}

    \vspace{1em}
    \begin{minipage}[t]{0.5\textwidth}
        \centering
        \includegraphics[width=\textwidth]{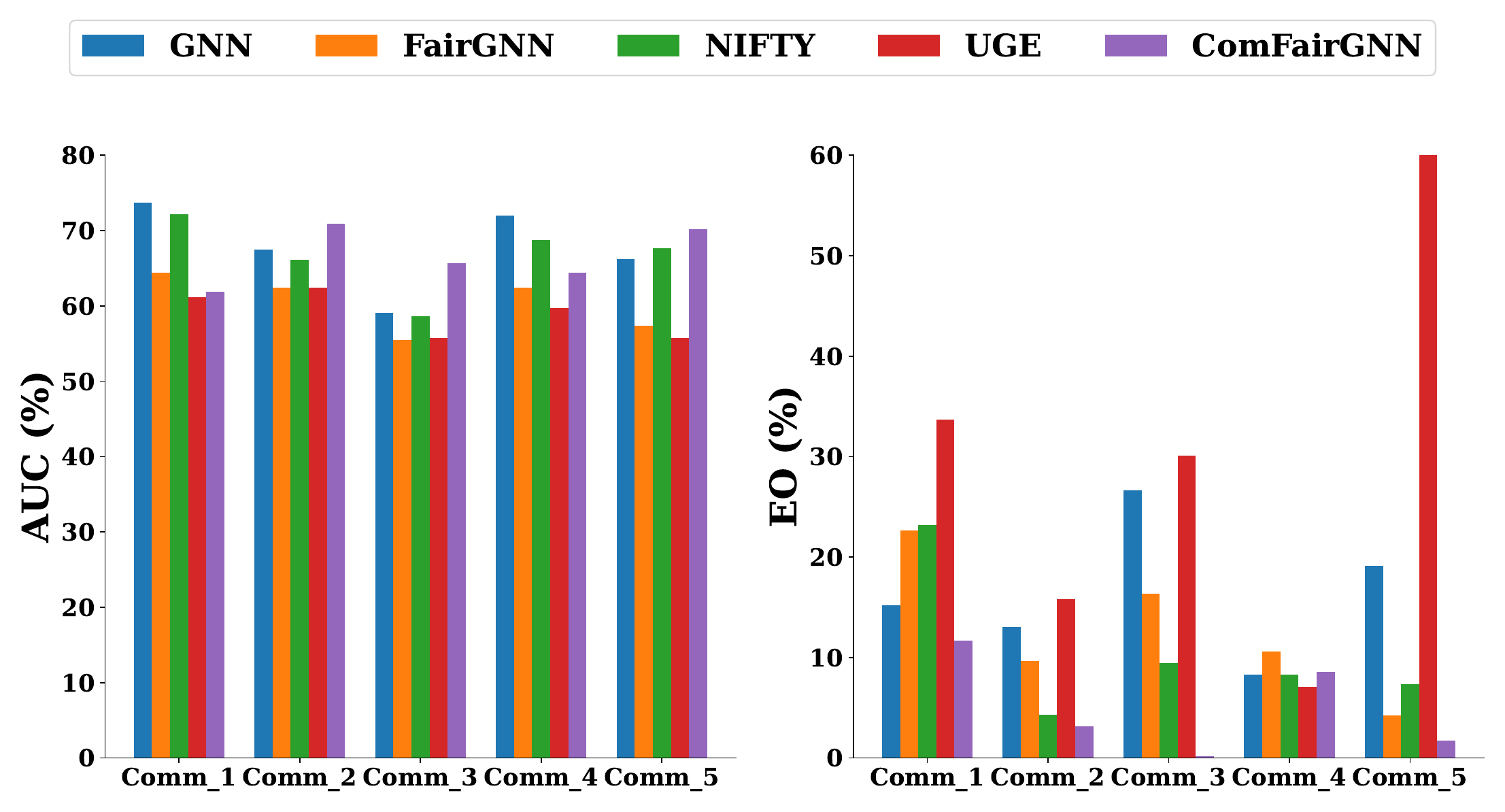}
        \caption*{(c) Performance on Pokec-n communities}
        \label{fig:dataset3}
    \end{minipage}
    \hfill
    \begin{minipage}[t]{0.48\textwidth}
        \centering
        \includegraphics[width=\textwidth]{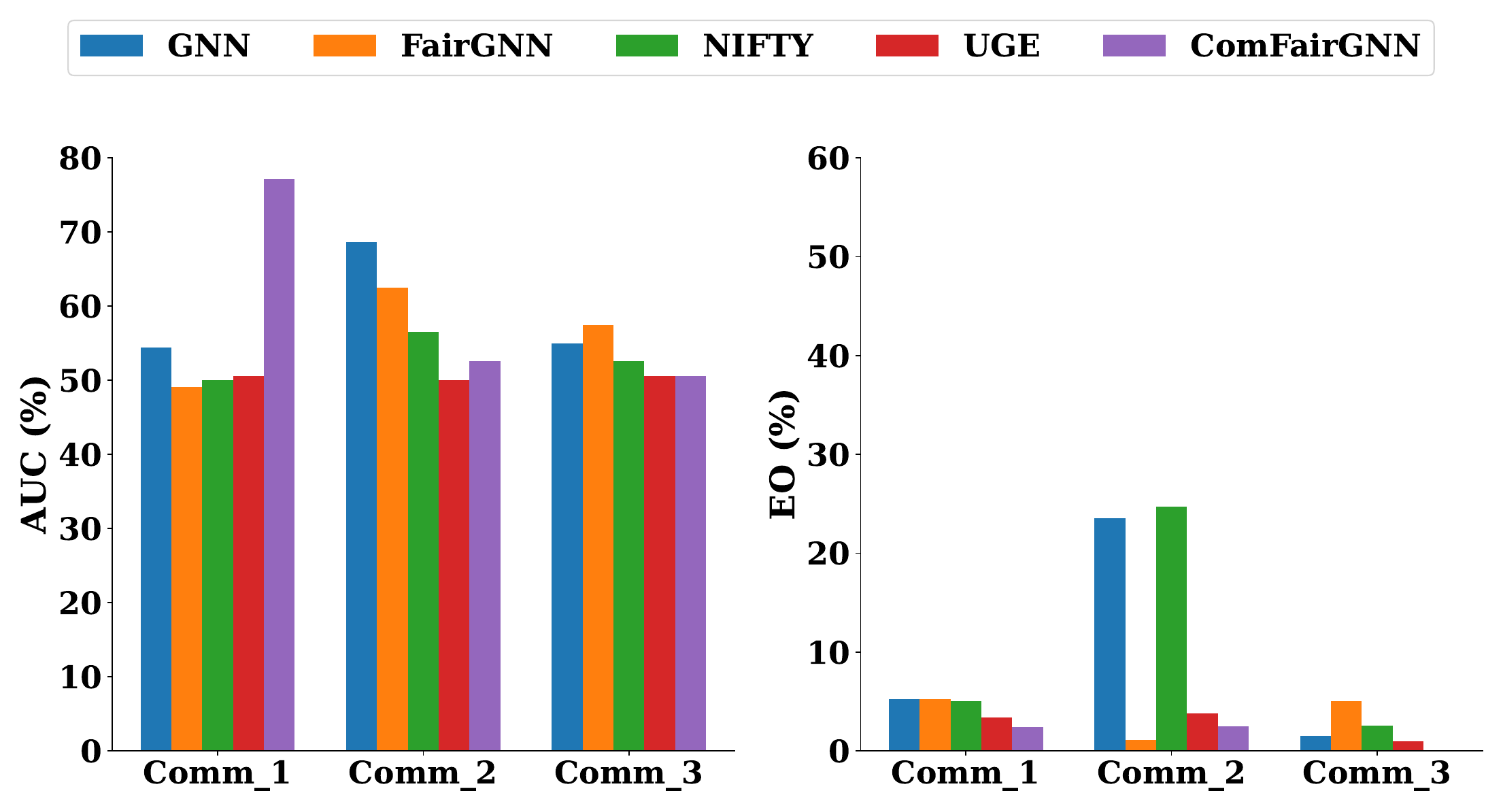}
        \caption*{(d) Performance on Fb communities}
        \label{fig:dataset4}
    \end{minipage}
    \caption{Performance comparison of different GNN debiasing methods across datasets and communities. (a) Graph-level performance across different datasets (Pokec-z, Pokec-n, Facebook). (b, c, and d) Community-wise performance breakdown on Pokec-z, Pokec-n and Facebook datasets. Showing AUC(\%) and EO(\%) metrics.}
    \label{fig:comparison}
\end{figure*}

\noindent\textbf{Fairness Paradoxes:}
To address \textbf{RQ1}, we evaluate the effectiveness of current debiasing techniques for GNNs at the community level and compare the consistency of their debiasing performance when applied to the entire graph. We first measure the group fairness of the GCN model with the debiasing method applied to the entire original input graph $G$. Next, we measure the group fairness of the same GCN model with the debiasing method applied at the community level. We observe that the fairness performance across the communities is not as consistent as the fairness score reported for the entire graph across datasets for all baseline methods. For example, in figure~\ref{fig:comparison} \{b, c, d\}, we observe that the unfairness estimation for Pokec-z dataset using $\Delta\text{EO}$ is higher for communities \{2, 3, 4, 5\} compared to $\Delta\text{EO}$ for the entire dataset when FairGNN debiasing method is used. 


\noindent\textbf{Over-Simplification of Fairness Evaluation \textbf{RQ2}:} 
The absolute values in $\Delta\text{SP}$ and $\Delta\text{EO}$ metrics can mask local disparities across communities. While overall metrics might show moderate fairness when averaged, they obscure cases where a group experiences high bias in some communities and low bias in others, as shown in Figure~\ref{fig:comparison}. This averaging effect fails to capture the actual fairness experienced by individuals in different parts of the graph.

\noindent\textbf{Evaluation and Main Results (RQ3)} As GNNs learn by neighborhood aggregation section~\ref{sec:gnn}, we evaluate fairness within different subgraphs or communities within the larger graph. This can help identify and address local disparities more effectively. We select $S_0$ and $S_1$ nodes within the community to evaluate the model's disparity locally, which may present more prominent biases. We select top and bottom $15$ sample homophily neighborhood distribution ratio section~\ref{sec:homophily_ratio}, from each community to formulate the representative coreset nodes from the training set during training. These nodes are sufficient to cover the local contextual structures where biases are rooted. As shown in Table ~\ref{tab:results}, ComFairGNN has both the best overall classification performance than the baselines methods. In terms of fairness, ComFairGNN respectively reduces $\Delta\text{SP}$ and $\Delta\text{EO}$ by 4.9\% and 4.1\%, compared with the best performed baseline on both Pokec-z and Pokec-n datasets. Additionally, as shown in Figure~\ref{fig:comparison}, this pattern is consistent across all communities in the datasets. Hence, the fairness improvement achieved by ComFairGNN aligning with our motivation.





\begin{figure}[t]
    \centering
    \includegraphics[width=0.35\textwidth]{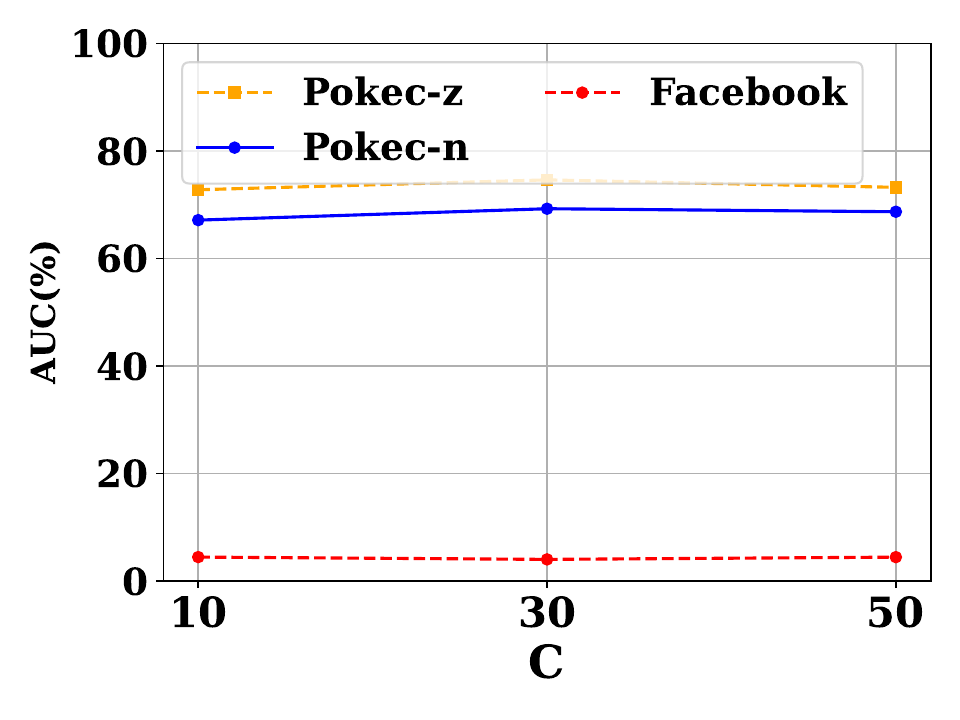}
    \hspace{0.5cm}
    \includegraphics[width=0.35\textwidth]{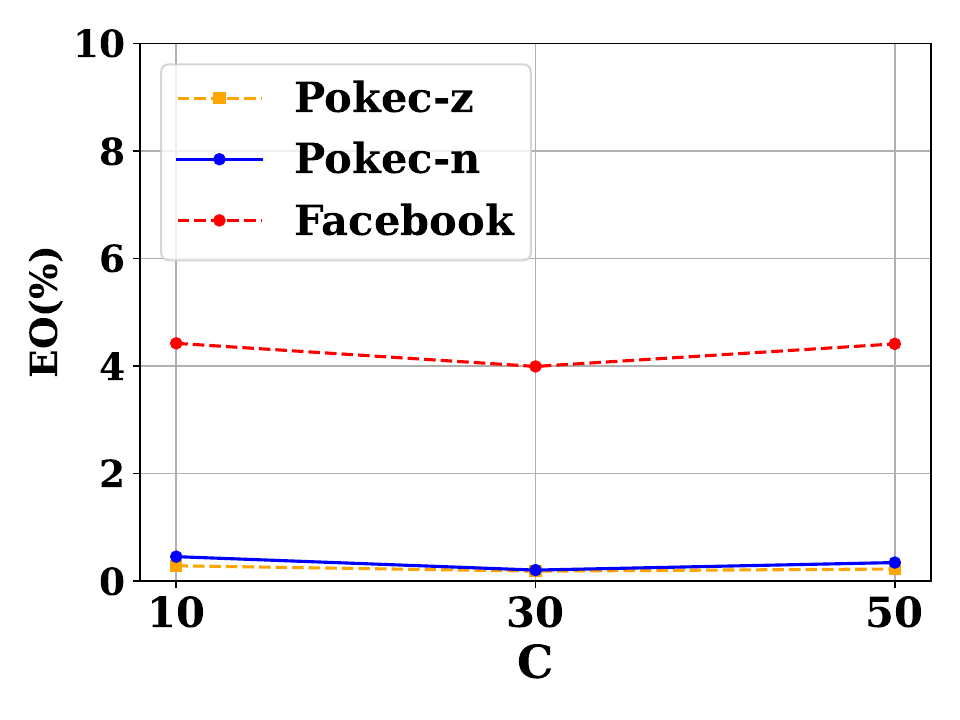}
    \caption{AUC and EO Performance Comparison Across Different Coreset Sizes $C$ }
    \label{fig:C_abilatioin}
\end{figure}

\noindent\textbf{Ablation Study:}
To validate ComFairGNN's, we investigate different coreset sizes of $C$ hyperparameter. As shown in figure~\ref{fig:C_abilatioin} $C = 30$ sample size relatively performs better.

%% file: body/related_work.tex
\section{Related Work}

\subsection{Fairness in Graph}
Research in Graph Neural Networks (GNNs) fairness has produced several approaches to measure and mitigate bias. Recent works have primarily focused on group fairness, ensuring models don't discriminate against certain groups. EDITS \cite{dong2022edits} modifies the adjacency matrix and node attributes to reduce the Wasserstein distance between different groups. FairGNN \cite{dai2021say} incorporates fairness regularization to ensure equitable treatment in both representations and predictions. UGE \cite{wang2022unbiased} learns unbiased node representations by eliminating sensitive attribute influences, while NIFTY \cite{agarwal2021unified} enhances counterfactual fairness through triplet-based objectives and layer-wise weight normalization. DegFairGNN \cite{liu2023generalized} addresses fairness for nodes with varying degrees within groups. However, this approach is computationally expensive for large graphs. Additionally, like other existing methods, DegFairGNN has not fully addressed the inherent local structural bias in graphs.


$$\mathcal{L}_{\text{total}} = \mathcal{L}_{\text{task}} + \lambda \times \mathcal{L}_{\text{fair}}$$



%% file: body/conclusion.tex
\section{Conclusion}
This paper examines fairness paradoxes in graph debiasing methods at the community level. By analyzing structural communities, we reveal fairness inconsistencies across demographic subgroups that are masked when using absolute measures of $\Delta\text{SP}$ and $\Delta\text{EO}$. To address community-level bias in GNN neighborhood aggregation, we propose ComFairGNN, a fairness-aware framework that modulates representative coreset nodes through embedding similarity contrast. Experiments on three benchmark datasets demonstrate that ComFairGNN effectively improves both accuracy and fairness metrics.
\section*{Acknowledgments}
This work is partially supported by the National Science Foundation under Grant No. 2152117 and NSF-CAREER 2237831. Any opinions, findings, conclusions, or recommendations expressed in this material are those of the author(s) and do not necessarily reflect the views of the National Science Foundation.